\documentclass[sigconf]{acmart}

\AtBeginDocument{%
  }

\setcopyright{acmlicensed}
\copyrightyear{2024}
\acmYear{2024}
\setcopyright{acmlicensed}\acmConference[DAC '24]{61st ACM/IEEE Design Automation Conference}{June 23--27, 2024}{San Francisco, CA, USA}
\acmBooktitle{61st ACM/IEEE Design Automation Conference (DAC '24), June 23--27, 2024, San Francisco, CA, USA}
\acmDOI{10.1145/3649329.3658498}
\acmISBN{979-8-4007-0601-1/24/06}
\usepackage{bm}
\usepackage{algorithmic}
\usepackage{algorithm}
\usepackage{enumitem}

\begin{document}

\title{APTQ: Attention-aware Post-Training Mixed-Precision Quantization for Large Language Models}

\author{Ziyi Guan$^{1,2*}$, Hantao Huang$^{1*}$, Yupeng Su$^1$, Hong Huang$^1$, Ngai Wong$^2$, Hao Yu$^1$}
\thanks{$^*$Equal contribution.}
% \authornote{ZG and HH contributed equally to this research.}
\affiliation{%
 \institution{School of Microelectronics, Southern University of Science and Technology, Shen Zhen, China$^1$ \\ Department of Electrical and Electronic Engineering, University of Hong Kong, Hong Kong, China$^2$}
 \country{} 
}
% \email{yuh3@sustech.edu.cn}

\renewcommand{\shortauthors}{Ziyi Guan et al.}
\begin{abstract}
Large Language Models (LLMs) have greatly advanced the natural language processing paradigm. However, the high computational load and huge model sizes pose a grand challenge for deployment on edge devices. To this end, we propose APTQ (Attention-aware Post-Training Mixed-Precision Quantization) for LLMs, which considers not only the second-order information of each layer's weights, but also, for the first time, the nonlinear effect of attention outputs on the entire model. We leverage the Hessian trace as a sensitivity metric for mixed-precision quantization, ensuring an informed precision reduction that retains model performance. Experiments show APTQ surpasses previous quantization methods, achieving an average of 4 bit width a 5.22 perplexity nearly equivalent to full precision in the C4 dataset. In addition, APTQ attains state-of-the-art zero-shot accuracy of 68.24\% and 70.48\% at an average bitwidth of 3.8 in LLaMa-7B and LLaMa-13B, respectively, demonstrating its effectiveness to produce high-quality quantized LLMs.
\end{abstract}

%    <concept>
%        <concept_id>10010147.10010257.10010293.10010307</concept_id>
%        <concept_desc>Computing methodologies~Learning linear models</concept_desc>
%        <concept_significance>300</concept_significance>
%        </concept>
%  </ccs2012>

% \ccsdesc[300]{Computing methodologies~Learning linear models}




\maketitle



%\maketitle

\section{Introduction}
\label{sec: intro}

%In the expanding realm of artificial intelligence, the prominence of Large Language Models (LLMs) such as ChatGPT~\cite{ouyang2022training}, OPT~\cite{zhang2022opt}, and LLaMA~\cite{touvron2023llama}, have surged, with their unparalleled prowess in deciphering and generating human language. Their utility spans an extensive array of applications, underscoring their critical role in advancing natural language processing. Yet, the deployment of these behemoth models is hampered by their exorbitant computational demands and memory footprints, confining their operation to high-end GPU farms and precluding their broader application. Acknowledging this barrier, the discourse in model compression has honed in on two prominent strategies: Quantization-Aware Training (QAT) and Post-Training Quantization (PTQ), the latter gaining traction for LLMs given the impracticability of QAT for models teeming with billions of parameters due to the excessive training overheads involved. OPT~\cite{zhang2022opt}

Large Language Models (LLMs), such as ChatGPT~\cite{ouyang2022training}, OPT~\cite{zhang2022opt}, LLaMA~\cite{touvron2023llama}, etc., exhibit impressive performance across various tasks. However, deploying these models on edge devices is challenging due to their exorbitant computational demands and memory footprints. Existing model compression solutions such as pruning~\cite{carreira2018learning} and neural architecture search~\cite{chen2021progressive} often require model retraining, which is extremely time-consuming and expensive for billion-parameter models. Recently, post-training quantization (PTQ) methods, such as GPTQ~\cite{frantar-gptq}, have been proposed and achieved relatively high accuracy without retraining. However, GPTQ only considers the weight quantization strategy in the scope of a single layer as an optimization problem to minimize 
$||\bm{W}\bm{X} - \bm{\hat{W}}\bm{X}||_2^2$, with $\bm{W}$, $\bm{\hat{W}}$ and $\bm{X}$ representing float weights, quantized weights and inputs, respectively. This simplification fails to consider the complex and nonlinear effects such as softmax in the attention computation, and leads to a sub-optimal solution. 
%However, GPTQ primarily focuses on the weight calculations within the feed-forward layers, engaging in a rather simplistic treatment of the weight matrix multiplications ($WX$).
%Such simplification becomes an even significant limitation when moving to even lower bit (2bit).

%Conventional strategies aimed at model compression, like the GPTQ~\cite{frantar-gptq} method, have indeed made strides towards mitigating this issue. Yet, GPTQ primarily focuses on the weight calculations within the feed-forward layers, engaging in a rather simplistic treatment of the weight matrix multiplications (WX). This simplification becomes a significant limitation when addressing the quantization of the q, k, and v projection layers within the attention mechanism. GPTQ's oversight of the updates in $W_q, W_k, W_v$ during quantization leads to a loss of critical information, which is instrumental for the attention mechanism's functionality within Transformer-based models.

%To bridge this gap, our paper presents the Attention-aware Post-Training Quantization (APTQ) technique, which is designed to remedy the deficiencies observed in GPTQ. APTQ introduces an innovative approach that accounts for the gradients of $W_q, W_k, W_v$, ensuring that the quantization process thoroughly considers the weight updates in the attention layers of the Transformer architecture. By doing so, APTQ significantly reduces the quantization error in these crucial components, thereby preserving the model's integrity during compression.

To achieve lower bitwidths without sacrificing the accuracy on edge devices, this paper presents an Attention-aware Post-Training Mixed-Precision Quantization (APTQ) technique, which is designed to consider the quantization optimization problem within the scope of the attention block including the nonlinear softmax operation. Specifically, APTQ utilizes gradients derived from the attention output and develops a second-order Hessian optimization strategy to quantize the weights. By doing so, APTQ significantly reduces the quantization error in these crucial components, thereby preserving the model's integrity throughout compression.

Furthermore, APTQ proposes a novel Hessian trace-based quantization sensitivity metric to implement mixed-precision quantization to further compress LLM models. This approach judiciously applies varying bitwidths across the model parameters to fit the limited memory size on edge devices with balanced size and accuracy. As a result, APTQ constitutes a mixed-precision 2/4-bit hybrid scheme with performance comparable to a uniform 4-bit representation. In particular, APTQ produces a compressed model close to its full-precision counterpart, and outperforming the GPTQ method especially in the realm of ultra-low-bit quantization scenarios. Through comprehensive experiments on the LLaMA-7B and LLaMA-13B models \cite{touvron2023llama}, the effectiveness of APTQ is validated on both perplexity and zero-shot performance, thus entailing a viable solution for the deployment of LLMs on edge devices.

%Moreover, APTQ utilizes a novel Hessian-based metric to implement mixed-precision quantization. This approach judiciously applies varying bit-widths within the model's parameters, balancing performance and efficiency. As a result, APTQ is not only able to achieve quantization levels that are close to uniform 4-bit representations but does so with the added benefit of a 2/4-bit hybrid scheme. This method yields a compressed model whose performance closely approximates that of the full-precision counterpart, outperforming the GPTQ method, especially in the realm of ultra-low-bit quantization scenarios. Through comprehensive experiments, such as those conducted on the LLaMA-7B model, APTQ's effectiveness is validated, presenting a viable solution for the deployment of LLMs in resource-constrained environments.
The main contributions of this paper are threefold:
\begin{itemize}[noitemsep,topsep=0pt]
   \item This is the first work to quantize LLMs by integrating
 the attention-based gradients with second-order Hessian optimization, leading to a nuanced update mechanism that enhances the precision throughout the quantization process.
\item An innovative Hessian trace-driven mixed-precision quantization scheme is proposed that judiciously allocates high/low bitwidths across different layers based on their sensitivity, optimizing model performance while maintaining efficiency.
\item Through extensive experimentation on the LLaMa models, APTQ not only achieves state-of-the-art (SOTA) results on the C4 dataset \cite{raffel2020exploring} but also attains near full-precision perplexity at an average quantization of 4 bits. %In zero-shot tasks, HAMQ also demonstrates superior performance compared to the majority of existing quantization approaches.
In zero-shot tasks, APTQ also 
demonstrates superior performance compared to the SOTA approaches.
\end{itemize}
\vspace{-5pt}

\section{Related Work}
\label{sec: related work}
% Network quantization, which involves converting the model's weights, activations, and gradients from floating-point to fixed-point integers or other discrete forms has become a widely adopted model compression technique. It significantly reduces the storage and computational overhead of the model. Quantization methods can be broadly categorized into two main classes: Quantization-Aware Training (QAT) and Post-Training Quantization (PTQ).
% \subsection{LLM Quantization}
%In LLM, quantization is regarded as an essential technique since it can significantly reduces the storage and computational overhead of the model. Quantization methods can be broadly categorized into two main classes: Quantization-Aware Training (QAT) and Post-Training Quantization (PTQ).
% LLM.int8() employs mixed-precision decomposition to fix the problem caused by activation outliers and vector-wise quantization for efficient inference. NuQmm specializes in highly efficient GPU kernels designed for a quantization scheme based on binary coding. 
%AWQ and OWQ both focus on the impact of activation outliers on weights. AWQ takes into consideration the importance of weight channels corresponding to larger activation magnitudes, whereas OWQ employs mixed-precision quantization to bestow higher precision upon weights that are susceptible to quantization-induced errors due to activation outliers.
To deploy large models on edge devices, quantization is a versatile technique for reducing model size and computation. Quantization-Aware Training (QAT) is known to be effective by integrating the quantization process into the training process. 
A representative work is LLM-QAT~\cite{liu2023llm-qat}, which proposes data-free distillation. However, this method introduces new trainable parameters, necessitates high-end GPU computational resources, and incurs a large time consumption. In contrast, Post-Training Quantization (PTQ) employs moderate resources to quantize pre-trained models without model retraining. Recent work, such as SpQR~\cite{dettmers2023spqr} and SqueezeLLM~\cite{kim2023squeezellm}, compress most weights to 4 bits but maintain outlier weights at 16 bits, which complicates the inference process with both 4-bit and 16-bit inference. 
% Recent work, such as SpQR~\cite{dettmers2023spqr} and 

% ZeroQuant~\cite{yao2022zeroquant} integrates a fine-grained and hardware-friendly quantization scheme that can be applied to weights and activations. 
SmoothQuant~\cite{xiao2023smoothquant} introduces a per-channel scaling transformation that effectively smooths the magnitudes to address the challenge of quantizing activations.
% RPTQ~\cite{yuan2023rptq} takes a unique clustering approach to organize channels into clusters for quantization, effectively mitigating the discrepancies in channel ranges. 
GPTQ~\cite{frantar-gptq} and OBQ \cite{frantar2022obq} introduce an innovative weight quantization method based on approximate second-order information, ensuring high accuracy and efficiency in the quantization process. Our work shares the same ethos as GPTQ but additionally considers the softmax and matmul operations within the attention computation to formulate the quantization problem, resulting in improved accuracy.

Mixed-precision quantization offers 
a trade-off strategy for edge devices to maintain the accuracy with minimized model size.
%significant benefits for Large Language Models (LLMs) on edge devices, reducing model size and minimizing storage and bandwidth requirements.
Existing works usually define some metrics to determine the quantization sensitivity of each layer.
One representative work is HAWQ-V2 \cite{dong2020hawq}, which adopts Hessian trace for CNN layer sensitivity assessment and utilizes the Hutchinson algorithm to approximately estimate the Hessian trace. Our APTQ method also employs Hessian trace for sensitivity but adopts the Levenberg-Marquardt approximation \cite{lecun1989obd} to directly calculate the Hessian trace with respect to the attention output, which is also an extension of GPTQ \cite{frantar-gptq} by further considering the nonlinear operation (softmax) and matmul in the attention output.  
Another close related work is PB-LLM \cite{shang2023pb}, which adopts a mixed 1-bit and fp-16 (half floating point) precision based on the Hessian values.
%Following a different approach, PB-LLM (\cite{shang2023pb}) employs a mixed-precision strategy combining full precision with 2-bit quantization, basing its decisions on calculated magnitudes or direct Hessian values, and developing its methodology on the foundation of GPTQ. 
Extreme low-bit quantization (1bit) is challenging for the accuracy. 
However, our APTQ method opts for a 2-bit and 4-bit mixed-precision quantization offering a better accuracy with the same model size comparing to PB-LLM.  The effectiveness of this strategy is demonstrated in Section \ref{sec: exp}, where our method shows superior performance in terms of efficiency and model compression when compared to PB-LLM.

\section{Algorithm}
\label{sec: algo}
% HHT version
This section starts with the preliminaries to outline the evolution of quantization techniques from optimal brain quantization (OBQ)~\cite{frantar2022obq} to our proposed Hessian-attention-based quantization. We then propose an Attention-aware Post-Training Mixed-Precision Quantization, APTQ, to further compress the LLMs. 

% choice 2: In pursuing efficiency enhancements for large neural networks, particularly models like Generative Pre-trained Transformers (GPT), quantization stands out as an essential technique. It serves to optimize storage and computational resources, with minimal accuracy loss. This necessity is especially pronounced in expansive models where the layer-wise quantization approach is often employed. We will trace the development of this technique from its inception in OBQ and GPTQ to its refined implementation in APTQ.
%choice 3: In the quest to enhance the efficiency of neural networks, particularly large models like ChatGPT, quantization emerges as a pivotal technique. It aims to reduce the storage and computational demands without significant loss of accuracy. Herein, we explore this problem's formulation through the lens of two leading approaches, OBQ and GPTQ, before introducing our novel APTQ algorithm.

\subsection{Preliminaries}
% OPQ --> GPTQ ---> APTQ 
% key point is the link to the problem
\textbf{General Quantization Framework.}
Quantization aims to reduce weight precision in neural networks, thus conserving computational resources. The general goal is to find a quantized weight matrix $\bm{\hat{W}}$ that approximates full precision output, minimizing the squared error. This process can be formally expressed as:
\begin{equation}
\text{argmin}_{\bm{\hat{W}}} ||\bm{W}\bm{X} - \bm{\hat{W}}\bm{X}||_2^2.
\label{eq1}
\end{equation}
In this equation, $\bm{X}$ represents the input to the layer, and $\hat{W}$ denotes the quantized weight.

\noindent\textbf{Optimal Brain Quantization (OBQ).}
Optimal Brain Quantization (OBQ) \cite{frantar2022obq} is an innovative method that minimizes quantization errors by treating each neural network weight independently. The core of OBQ lies in iteratively quantizing each weight and adjusting the remaining unquantized weights to compensate for the quantization-induced errors. This approach is mathematically articulated as follows:
\begin{equation}
w_{q} = \text{argmin}_{w_{q}} \frac{\text{quant}(w_{q}) - w_{q}}{[H_{F}^{-1}]_{qq}},
\label{eq2}
\end{equation}
\begin{equation}
\delta_{F} = -\frac{w_{q} - \text{quant}(w_{q})}{[H_{F}^{-1}]_{qq}} \cdot (H_{F}^{-1})_{:,q},
\label{eq3}
\end{equation}
\begin{equation}
H_{-q}^{-1} = (H^{-1} - \frac{1}{[H^{-1}]_{qq}}H_{:,Q}^{-1}H_{q}^{-1:})_{-p}.
\label{eq4}%
\end{equation}%
The Hessian matrix $H_F = 2X_FX_F^T$ guides the selection of the quantization candidate $w_q$ from the full-precision weights $F$, and the update $\delta_F$ is calculated to minimize quantization error, as formalized in equations (\ref{eq2}), (\ref{eq3}) and (\ref{eq4}) with $\text{quant}(w)$ mapping weights to their nearest quantized values. Building upon OBQ, GPTQ \cite{frantar-gptq} extends the principles by adopting the fixed order weights update strategy and Cholesky reformulation to speed up the computation.

%\noindent\textbf{Generative Pre-trained Transformer Quantization (GPTQ)}
% GPTQ extends the principles of OBQ to enhance scalability for LLMs. It introduces a uniform method for quantizing weights across different layers:
% Building upon OBQ, GPTQ \cite{frantar-gptq} extends the principles to enhance scalability for LLMs. It proposes a uniform quantization order across all rows, maintaining the same set of unquantized weights $F$. 

%The GPTQ method is mathematically represented by:
%\begin{equation}
%\delta_{F} = -(w_{Q} - \text{quant}(w_{Q}))([H_{F}^{-1}]{QQ})^{-1}(H_{F}^{-1})_{:,Q}
%\label{eq4}
%\end{equation}
%\begin{equation}
%H_{-Q}^{-1} = (H^{-1} - H_{:,Q}^{-1} ([H^{-1}]_{QQ})^{-1}H_{Q}^{-1:})_{-Q}
%\label{eq5}
%\end{equation}https://www.overleaf.com/project/6532347a3ba75e19e19de29b

\subsection{Hessian-Attention-based Quantization}
While GPTQ effectively minimizes layer-specific quantization errors, it overlooks the intricate nonlinearities in attention mechanisms, leading to suboptimality. APTQ, by contrast, embraces a holistic quantization strategy, factoring in the entire attention block and its nonlinear dynamics, which sharpens the precision of the quantized model, particularly in low-bitwidth scenarios.

\begin{figure*}[t]
  \centering
  \includegraphics[width=\linewidth]{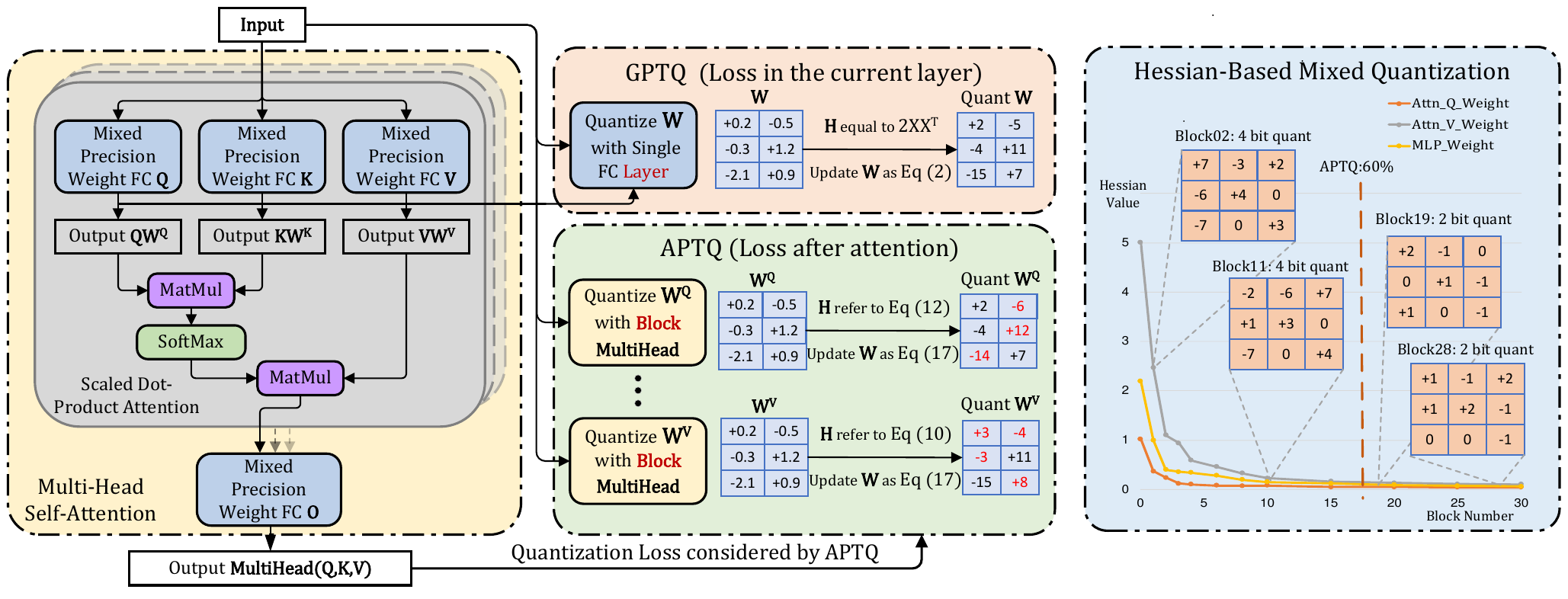}
  \caption{Overall architecture of APTQ (Attention-aware Post-Training Mixed-Precision Quantization): Unifying comprehensive transformer attention analysis with layer-specific Hessian trace quantization for enhanced model understanding.}
  \label{fig: overall}
\end{figure*}

As shown in Figure.\ref{fig: overall}, we present the advanced architecture of APTQ, demonstrating its comprehensive quantization strategy. Unlike GPTQ, which primarily processes loss in the current layer, APTQ integrates a full-scope analysis of the attention mechanism, including the $Q$, $K$, $V$, $O$ matrices, matmul and nonlinear activation layers such as softmax. This extensive approach not only focuses on the intricacies beyond simple weight matrix multiplication, but also significantly mitigates quantization errors, offering a robust solution in low-bitwidth quantization scenarios.

% textbf{(Left)} HAMQ vs. GPTQ: Enhanced attention layer processing, with focus on nonlinear information in Transformers. \textbf{(Right)} Distinct Hessian Matrix Update Methods in HAMQ: Gradient derivation for QKV attention layers.

\noindent\textbf{Objective Function}. At a macroscopic level, our methodology employs a layer-wise quantization approach to address the quantization reconstruction problem for each layer's weights. In the Transformer architecture, two main structural levels exist: the attention layers and the feed-forward layers. Specifically, in contrast to GPTQ, which treats each weight matrix as a linear layer and ignores the impact of other structures on the output, we treat all structures of the same layer as a whole, represented by the function $F$ standing for the attention output $\text{Multihead}(Q,K,V)$. We aim to reformulate Equation~(\ref{eq1}) and minimize the new squared error equation as follows:
\begin{equation}
  \text{argmin}_{\hat{W}} ||F(W)-F(\hat{W})||^2_2.
\label{eq: define}
\end{equation}
where $W$ remains constant and $\hat{W}$ is the quantized weights to be optimized.
%Assuming the pre-quantized weights are fixed ($W$ remains constant), we define the loss function as the squared Euclidean norm of the output error, which depends only on $\hat{W}$:
%\begin{displaymath}
%  E(\hat{W})=||F(W)-F(\hat{W})||^2_2
%\end{displaymath}
%The gradient and Hessian matrix of this 
The  Hessian matrix of this function is computed as:
%\begin{equation}
%  \nabla_{\hat{W}}(E)=-2\cdot[F(W)-F(\hat{W})] \cdot {F'(\hat{W})^T}
%\end{equation}
%\textcolor{red}{ADD SECOND ORDER EQUATION}
%The second order gradient can be computed as follows:
\begin{equation}
H_{\hat{W}} = 2 \cdot \left( F'(\hat{W}) \cdot F'(\hat{W})^T + [F(W) - F(\hat{W})] \cdot F''(\hat{W}) \right).
\label{eq:hessian_all_terms}
\end{equation}
This is the general expression of Hessian matrix.
%If $F(W)$ represents $\bm{WX}$, $F''(\hat{W})^T=0$, we get the GPTQ Hessian matrix $\bm{2XX^T}$.
%In our case, 
To ensure $H_{\hat{W}}$ is positive definite and invertible, we only retain the first-order derivative portion as the expression for the Hessian matrix, which is widely known as the Levenberg-Marquardt approximation~\cite{lecun1989obd}:
\begin{equation}
\label{sim_hessian}
  H_{\hat{W}}=2\cdot[F'(\hat{W})\cdot F'(\hat{W})^T].
\end{equation}
%The loss landscape of a well trained networks should be in the flat region, the second order of $F''(\hat{W})$ is thereby small enough to be ignored. 

\noindent\textbf{Derivatives for Different Quantization Layers. } The current problem is transformed into finding the partial derivative of $F(\hat{W})$ with respect to the weights $\hat{W}$. The $F(\hat{W})$ function is different for the Feed-Forward layers and Attention layers. 
In the Feed-Forward layer, the main structure is a linear fully connected layer.
%\begin{equation}
%  F(W,X)=WX 
%\end{equation}
The Hessian matrix is easily computed as $H_F = 2X_FX_F^T$, corresponding to the Hessian matrix form in the GPTQ method.

In the Attention layer, a multi-head mechanism is employed, where each attention head contains an Attention function:
\begin{equation}
F(W,X)=\mathrm{MultiHead}(Q,K,V).
\label{eq:multihead}
\end{equation}

The quantized weight matrices lead to different derivatives.
When quantizing the $W^O$ matrix, consider $W^Q$, $W^K$, $W^V$ as constants:
\begin{equation}
\label{dwo}
\frac{\partial F}{\partial W^O}=\text{Concat}(\text{head}_1,...,\text{head}_\text{H})^T\frac{\partial F}{\partial X}.
\end{equation}
When quantizing the $W^V$ matrix, consider $W^Q$, $W^K$, $W^O$ as constants:
\begin{equation}
\label{dwv}
\frac{\partial F}{\partial W^{V}}={M}^{T}\frac{\partial F}{\partial X}(W^{O})^{T}.
\end{equation}
Here, $M$ represents a matrix composed of $H$ heads losing $W_i^V$:
\begin{equation}
M_h=\mathrm{softmax}(\frac{QW_{h}^{Q}(W_{h}^{K})^{T}K^{T}}{\sqrt{d_{k}}})V, M = \left[M_1,\ldots,M_H\right].
\end{equation}
When quantizing $W^Q$ or $W^K$ matrices, consider the remaining three terms as constants:
\begin{equation}
\label{dwq}
\frac{\partial F}{\partial W^{Q}_h}=\frac{1}{\sqrt{d_k}} Q^{T}\frac{\partial F}{\partial N}\mathbb{P}_{h} ^{T}KW_{h}^{K},
\end{equation}
\begin{equation}
\label{dwk}
\frac{\partial F}{\partial W_h^K}=\frac{1}{\sqrt{d_k}} K^T\mathbb{P}_h\frac{\partial F}{\partial N}^T QW_h^Q.
\end{equation}
Here, $W_h$ represents the weight matrix in the $n$-th attention head, and $N$ and $\mathbb{P}_h$ are given by:
\begin{equation}
N_h=\frac{QW_h^Q(W_h^K)^TK^T}{\sqrt{d_k}},\ N=[N_1,\ldots,N_H],
\end{equation}
\begin{equation}
\quad \mathbb{P}_h=(\ldots,,E_{n\times n}^h,\ldots)_{n\times nH}.
\end{equation}%
After computing the gradients from equations (\ref{dwo}), (\ref{dwv}), (\ref{dwq}) and (\ref{dwk}), we can further get their second order gradients using equation (\ref{sim_hessian}) to obtain the corresponding Hessian matrix. Thus, referring to the optimization problem in equation~(\ref{eq: define}), combining the quantization techniques in equations~(\ref{eq2}), (\ref{eq3}), we derive the following formulas for updating weights in the context of attention mechanisms:

\begin{equation}
E = -\frac{w_{q} - \text{quant}(w_{q})}{([H_{\hat{W}}^{-1}]_{qq})},
\label{eq: our_error}
\end{equation}

\begin{equation}
\delta_{F} = {E} \cdot (H_{\hat{W}}^{-1})_{:,q}.
\label{eq: update_w}
\end{equation}
Here,${E}$ represents the quantization error, $w_q$ refers to the quantized weights of the current group. $\delta_{F}$ refers to the corresponding optimal updates for the remaining float weights (not yet quantized weights of the current layer). This principle is uniformly applicable to the quantization of $Q$ (query), $K$ (key), $V$ (value), and $O$ (output) weight matrices in attention mechanisms. By synthesizing these elements, we can effectively compute the second-order Hessian information relevant to the weights within the attention layers. This advanced computation aids in the update and optimization of weights, targeting the minimization of the original squared error as defined in equation (\ref{eq: define}). This approach facilitates the realization of quantized models with robust performance across different components of the attention mechanism. The comprehensive algorithm is detailed in Algorithm Box~\ref{alg:HAMQ}.

% , where ${E}^{V}$ is the quantization error in attention V matrix and $\delta_{F}^{V}$ denotes the greedy-optimal weight to quantize next. All in all, we can calculate the second-order hessian information of weight in the attention layers and update the weights to optimize the original squared error to get the high-accuracy quantized models.

% \begin{equation}
% \delta_{F} = -\frac{w_{q} - \text{quant}(w_{q})}{[H_{F}^{-1}]_{qq}} \cdot (H_{F}^{-1})_{:,q}
% \label{eq3}
% \end{equation}
% optimize the quantized weights

\begin{algorithm}[ht]
\small
\algsetup{linenosize=\small}
\caption{APTQ via Hessian-Attention-based Mixed-Precision Quantization}
\label{alg:HAMQ}
% Input
{\bf Input:} Pre-trained model weights $W$, blocksize $B$, Hessian matrix $H$, quantization function $\text{quant}$, Layer names $layerName$, Ratio of 4-bit in 2/4 mixed-precision $R$.
\begin{algorithmic}[1]
\STATE Initialize quantized weight matrix $Q \leftarrow 0_{d_{\text{row}} \times d_{\text{col}}}$.
\STATE Initialize block quantization error matrix $E \leftarrow 0_{d_{\text{row}} \times B}$.
\STATE \textbf{Step 1: 4-bit Hessian-Attention-Based Quantization}
\FOR{$i = 0, B, 2B, \ldots$}
    \FOR{$j = i, \ldots, i + B - 1$}
        \IF{\( \text{‘‘self\_attn.k\_proj’’} \) in \( \text{layerName} \)}
            \STATE $ H_{\hat{W}}^K = 2[\frac{\partial F}{\partial W^{K}} \cdot \frac{\partial F}{\partial W^{K}}^T]$ from Equation~(\ref{dwk})
            \STATE \( Q_{:,j}^K \leftarrow \text{quant}(W_{:,j}) \)
            \STATE \( E_{:,j-i}^K \leftarrow (W_{:,j}^K - Q_{:,j}^K) / [H_{\hat{W}}^{-1}]_{jj}^K \) based on Equation~(\ref{eq: our_error})
            \STATE \( W_{:,j:(i+B)}^K \leftarrow W_{:,j:(i+B)}^K - E_{:,j-i}^K \cdot (H_{\hat{W}}^{-1})_{:,j:(i+B)}^K \) based on Equation~(\ref{eq: update_w})
            \STATE For self\_attn.Q, V, and O projection layers, similar updates are applied
            \STATE Compute the average Hessian trace for each layer in block \( i:(i+B) \).
        \ENDIF
    \ENDFOR
\ENDFOR
\STATE \textbf{Step 2: Hessian-trace-based Mixed-Precision Quantization}
\STATE Calculate Hessian trace values for each layer, and order them from highest to lowest, starting with the previously established 4-bit quantization.
\STATE Determine the layers for mixed-precision quantization based on the computed Hessian trace values and $R$.
\FOR{each selected layer}
    \STATE Calibrate the bit allocation in line with each layer's Hessian trace sensitivity and $R$.
    \STATE Implement 2/4 bit mixed-precision quantization
\ENDFOR
\end{algorithmic}
{\bf Output}: The resulting quantized model weights $Q$ are characterized by scale, zero-point, and quantization error.
\end{algorithm}

\subsection{Hessian-Trace-based Mixed-Precision Quantization}
% The assessment of layer sensitivity through Hessian Trace is a pivotal indicator for quantization, originally introduced by \cite{dong2020hawq}.
As mentioned in Section~\ref{sec: related work}, the Hessian trace provides sensitivity information for implementing mixed-precision quantization. 
Figure~\ref{fig: overall} illustrates the APTQ method's allocation of 4-bit and 2-bit quantizations, utilizing average Hessian trace values as a measure of layer sensitivity. This approach diverges from the GPTQ method, which concentrates solely on the matrix multiplication within the current layer, while APTQ provides a comprehensive assessment of each layer's impact. 

By computing the average trace of the Hessian matrix, the method determines the appropriate level of precision for the quantization of each layer. Layers with higher Hessian Trace values, which exert a greater influence on the network's output, require higher bit precision to ensure the model's accuracy. 
%In contrast, layers with lower trace values can undergo more substantial quantization, enabling reductions in model size and enhancements in computational efficiency. This systematic approach, integral to the HAMQ methodology, ensures a balanced quantization of layers, optimizing performance and efficiency in tandem.
Utilizing this mixed-precision quantization scheme results in models with an average bit precision defined by the formula:
\begin{equation}
\text{average bits} = 4 \times R + 2 \times (1-R),
\label{eq: ratio}
\end{equation}
where $R$ denotes the proportion of weights quantized at 4 bits within the overall quantization process.
This formula is a pivotal aspect of the APTQ methodology, facilitating a dynamic adjustment that is particularly advantageous for deploying large language models on edge devices. The adaptability of $R$ allows the APTQ algorithm to allocate higher precision to layers with greater sensitivity, while applying more robust quantization to less sensitive layers. Consequently, this leads to a quantized model that achieves an optimal balance between performance and size to deploy on edge devices.

% The quantization process culminates in the calculation of the average bits for the quantized model. This is achieved using the formula \(\text{average bits} = 4 \times R + 2 \times (1-R)\), where \( R \) signifies the ratio of layers quantized at 4 bits to those at 2 bits. This formula represents a critical aspect of the HAMQ methodology, allowing for a dynamic balance between model accuracy and computational efficiency. By adjusting \( R \), the HAMQ algorithm flexibly allocates higher precision to layers with greater sensitivity (as indicated by their Hessian Trace values) while applying more aggressive quantization to less sensitive layers. This results in a quantized model that optimally balances performance and size, tailored to the specific needs and characteristics of the model.

Algorithm~\ref{alg:HAMQ} unfolds into two decisive steps aimed at enhancing model efficiency while preserving performance. Step 1 applies 4-bit quantization to the attention mechanism's $K$ (key) layer, guided by the Hessian matrix, $H_{\hat{W}}^K$, that entails the second-order derivative crucial for this optimization, as formulated in Equation~(\ref{dwk}). This step adjusts the precision of the $K$ layer's weights, considering the broader implications for the model's performance. The individual optimization of the $K$, $Q$, $V$, and $O$ layers is informed by their respective Hessian matrices, ensuring that quantization is precisely targeted to maintain the balance between efficiency and accuracy. In essence, Hessian-Attention-based quantization strategically refines weight precision within attention layers to maintain model accuracy without unnecessary computational burden.

In the algorithm's second phase, a mixed-precision quantization strategy is implemented, beginning with the calculation of Hessian trace values across the layers. These values are then ordered in a descending sequence, starting with the layers previously quantized at a 4-bit level. This ordering informs the selection of layers for subsequent mixed-precision quantization, which is performed in accordance with the computed Hessian trace values. This selective quantization process is designed to align closely with each layer's functional impact on the overall model, ensuring a quantization scheme that is both effective and efficient.

% This algorithmic framework synthesizes the concepts of HTMQ into a coherent sequence of steps that can be integrated into the larger context of the APTQ algorithm. It ensures that the quantization process is not just a blind application of reduced precision but a calculated optimization that preserves the delicate balance between the model's complexity and computational constraints.

\section{Experiment}
\label{sec: exp}
% \vspace{-0.2cm}
\subsection{Experiment Setup}
To evaluate APTQ's performance, we focus on two primary metrics: perplexity and zero-shot performance. The LLaMa family~\cite{touvron2023llama} serves as the foundation for our experiments, owing to its efficacy and critical influence in recent model advancements. To maintain consistency and comparability, our benchmarking procedures against GPTQ adhere to identical experimental configurations. Our calibration dataset encompasses 128 segments, each containing 2048 tokens randomly sampled from the C4 dataset. All experiments deploy a group size of 128 and are executed on a single NVIDIA A100 GPU of 80GB memory. 
%Note that the quantization parameters such as scale and zero value are float 16 (fp16). Therefore, the average bit width is slightly larger than 4 bit for 4-bit quantization. 
Our APTQ is applied directly to the pre-trained model (post-training quantization). The evaluation of zero-shot performance is conducted using the EleutherAI/lm-evaluation-harness~\cite{leo_gao_2022_7413426}.  
Note that we use the format APTQ-R to represent the mixed precision (2/4-bit) setting, with $R$ represents the percentage of 4-bit weights as discussed in Equation (\ref{eq: ratio}).

% In addition to the standard 4-bit quantization comparison with GPTQ, we also implement our innovative Hessian trace-based mixed-precision quantization to contrast with other mixed-precision PTQ methods. This process ranks each layer by the average Hessian trace relative to its parameter count. Layers with higher average Hessian traces receive more bits, indicating their greater importance, while those with lower traces are allocated lower bits, optimizing the balance between model precision and size efficiency.

%\vspace{-1.0cm}
\begin{table}[t]
\small
\centering
\setlength{\abovecaptionskip}{0pt}
\setlength{\belowcaptionskip}{-0.5cm}
\caption{Comparison of Perplexity of Quantized LLaMa Models on C4 and WikiText-2 Datasets.}
\begin{tabular}{lccc}
\toprule
\textbf{Method} & \textbf{Avg bit} & \textbf{C4 $\downarrow$} & \textbf{WikiText-2 $\downarrow$} \\
\hline
LLaMa-7B & 16 & 5.22 & 5.68 \\
GPTQ \cite{frantar-gptq} & 4.0 & 5.62 & 8.14 \\
OWQ \cite{lee2023owq} & 4.01 & 5.56 & 7.15 \\
LLM-QAT \cite{liu2023llm-qat} & 4.0 & 7.40 & 10.90 \\
PB-LLM-20\% \cite{shang2023pb} & 3.4 & 20.61 & 17.19 \\
\textbf{APTQ} & \textbf{4.0} & \textbf{5.23} & \textbf{6.45} \\
\textbf{APTQ-75}\% & 3.5 & 5.54 & 6.54 \\
\textbf{APTQ-50}\% & 3.0 & 6.24 & 6.76 \\
 \bottomrule
\end{tabular}
\label{tab:perplexity_comparison}
\vspace{-0.3cm}
\end{table}
% \vspace{-0.6cm}

\subsection{Evaluation of Perplexity performance}
% \vspace{-0.2cm}
We assess the the performance of APTQ using the C4~\cite{raffel2020exploring} and WikiText-2~\cite{merity2016pointer} benchmarks. We compare APTQ against three established PTQ methods: GPTQ~\cite{frantar-gptq}, OWQ~\cite{lee2023owq}, and PB-LLM~\cite{shang2023pb}. Notably, OWQ and PB-LLM extend upon GPTQ, with PB-LLM incorporating mixed-precision quantization. To ensure a balanced comparison, all methods are evaluated on a standardized platform. Moreover, we benchmark APTQ's performance with the leading QAT approach, LLM-QAT. Table~\ref{tab:perplexity_comparison} reveals that APTQ, at an average 4 bit, closely matches the full-precision model and attains SOTA performance on the C4 dataset, showing only a 0.01-point increase in perplexity. Remarkably, even with average bit rates reduced to 3.5 and 3.0, APTQ's perplexity remains comparable to that of GPTQ's 4-bit model. This evidence of APTQ's stability at low bit rates positions it as a potent tool for optimizing the quantization and deployment of large-scale language models like LLaMa-7B.

To substantiate the robustness and broad applicability of the Hessian trace-based mixed-precision quantization posited in our study, we conducted a comparative analysis of various 4-bit utilization levels of APTQ against other prevalent PTQ methods applied to the LLaMa-7B model on the C4 dataset. The APTQ model, quantized at an average of 4 bit, not only approaches the full-precision model's perplexity but also outperforms all other PTQ approaches at a reduced precision of 3.5 bits. Impressively, configurations below 3 bits still surpass the 4-bit LLM-QAT baseline, underscoring APTQ's efficacy. These results unequivocally demonstrate the superior performance of APTQ, leveraging Hessian trace-driven precision allocation to optimize quantization outcomes.

Figure~\ref{fig: compare_ratio} visually summarizes our findings. It presents the comparative perplexity results of the LLaMa-7B model using APTQ at various bit utilization ratios when benchmarked against other PTQ and QAT methods on the C4 dataset. As depicted in the figure, the APTQ model consistently maintains competitive performance, even at significantly reduced bit rates. This graphical representation reinforces the effectiveness of the Hessian trace-based mixed-precision approach we advocate in this study, illustrating its potential for resource-efficient large model deployment.
% Comparative perplexity results of LLaMa-7B using HAMQ at various bit utilization ratios against other PTQ and QAT methods on the C4 dataset.
% The HAMQ model maintains competitive performance, even at significantly reduced bit rates, demonstrating the effectiveness of the Hessian trace-based mixed-precision approach.
% Resource-Efficient LLaMa-7B Quantization: HAMQ's Competitive Performance at Reduced Bit Rates

\begin{table*}[t]
\small
\renewcommand{\arraystretch}{1.32}
\setlength{\belowcaptionskip}{-0.3cm}
\centering
\caption{Zero-shot accuracy of quantized LLaMa models on common sense reasoning tasks.}
\label{tab:zero-shot-accuracy}
\setlength\tabcolsep{3pt} % Reduce the padding between columns
\scalebox{0.9}{
\begin{tabular}{@{}l|c|cccccc|ccccccr@{}}
\toprule
Model & \multicolumn{1}{c}{} & \multicolumn{6}{c}{LLaMa-7B} & \multicolumn{6}{c}{LLaMa-13B} \\
\cmidrule{1-14}
Method & Avg bit & PIQA & Hellaswag & Arc-E & Arc-C & WinoGrande & $\overline{Acc}\%\uparrow$ & PIQA & Hellaswag & Arc-E & Arc-C & WinoGrande & $\overline{Acc}\%\uparrow$ \\
\midrule
FP16 & 16 & 79.2 & 76.2 & 72.8 & 44.7 & 69.9 & \textbf{68.56} & 80.3 & 79.0 & 74.8 & 47.9 & 72.7 & \textbf{70.94} \\
RTN \cite{liu2023llm-qat} & 4.0 & 77.3 & 72.7 & 68.8 & 43.1 & 66.9 & 65.76 & 79.1 & 76.8 & 72.6 & 46.5 & 70.5 & 69.10 \\
SmoothQuant \cite{xiao2023smoothquant} & 4.0 & 76.4 & 68.1 & 67.3 & 39.6 & 66.0 & 63.48 & 77.9 & 74.2 & 76.3 & 45.5 & 69.7 & 68.72 \\
% QLLM & 6 & 77.26 & 71.4 & 52.02 & 41.04 & 65.19 & 61.38 & 77.26 & 71.4 & 52.02 & 41.04 & 65.19 & 61.38 \\
FPQ \cite{liu2023llm} & 4.0 & 77.8 & 75.0 & 72.4 & 41.7 & 69.0 & 66.60 & 79.4 & 77.7 & 72.8 & 47.3 & 71.5 & 69.74 \\
LLM-QAT \cite{liu2023llm-qat} & 4.0 & 78.3 & 74.0 & 70.0 & 41.7 & 69.0 & 66.60 & 79.4 & 77.7 & 72.8 & 47.3 & 71.5 & 69.74 \\
GPTQ \cite{frantar-gptq} & 4.0 & 76.0 & 69.4 & 66.9 & 43.0 & 66.7 & 64.40 & 79.8 & 77.7 & 73.2 & 45.9 & 72.6 & 69.84 \\
PB-LLM 30\% \cite{shang2023pb} & 4.1 & 78 & 74.3 & 69.0 & 42.3 & 69.7 & 66.66 & - & - & - & - & - & - \\
PB-LLM 10\% \cite{shang2023pb} & 2.7 & 67.8 & 68.1 & 58.7 & 39.6 & 67.4 & 60.32 & - & - & - & - & - & - \\
\textbf{APTQ} & 4.0 & 78.6 & 75.7 & 72.4 & 44.4 & 69.3 & 68.08 & 79.9 & 78.8 & 73.9 & 47.0 & 72.1 & 70.34 \\
\textbf{APTQ-90\%} & \textbf{3.8} & 78.8 & 75.9 & 73.6 & 43.5 & 69.4 & \textbf{68.24} & 79.4 & 78.8 & 73.8 & 47.8 & 72.6 & \textbf{70.48}\\
APTQ-80\% & 3.6 & 78.0 & 75.3 & 70.2 & 43.7 & 69.5 & 67.34 & 79.5 & 78.2 & 72.8 & 46.5 & 72.6 & 69.92 \\
APTQ-75\% & 3.5 & 77.5 & 74.5 & 68.7 & 44.2 & 70.2 & 67.02 & 79.3 & 77.6 & 71.8 & 46.1 & 73.2 & 69.60 \\
APTQ-70\% & 3.4 & 77.6 & 73.4 & 66.9 & 41.3 & 68.9 & 65.62 & 78.3 & 77.5 & 71.4 & 46.3 & 72.5 & 69.20 \\
APTQ-60\% & 3.2 & 76.8 & 72.1 & 63.1 & 39.3 & 69.5 & 64.16 & 78.6 & 74.2 & 69.5 & 44.2 & 69.5 & 67.20 \\
APTQ-50\% & 3.0 & 74.5 & 68.3 & 57.9 & 36.4 & 65.3 & 60.48 & 74.4 & 71.2 & 64.1 & 41.0 & 68.0 & 63.74 \\
\bottomrule
\end{tabular}
}
\end{table*}

\vspace{-0.2cm}
\subsection{Evaluation of Zero-shot performance}
In the evaluation of zero-shot performance, we extend our investigation to a suite of challenging zero-shot language tasks. These tasks, which span Predictive Question Answering (PIQA), Hellaswag, ARC-Easy (Arc-E), ARC-Challenge (Arc-C), and WinoGrande, serve as a benchmark for common sense reasoning in machine comprehension. We compare the proposed APTQ method on LLaMa-7B and LLaMa-13B with other advanced quantization techniques including round-to-nearest (RTN), SmoothQuant~\cite{xiao2023smoothquant}, FPQ~\cite{liu2023llm}, LLM-QAT~\cite{liu2023llm-qat}, and GPTQ~\cite{frantar-gptq}.
\begin{figure}[t]
\setlength{\abovecaptionskip}{0cm}
\setlength{\belowcaptionskip}{-0.7cm}
\centering
\includegraphics[scale = 0.50]{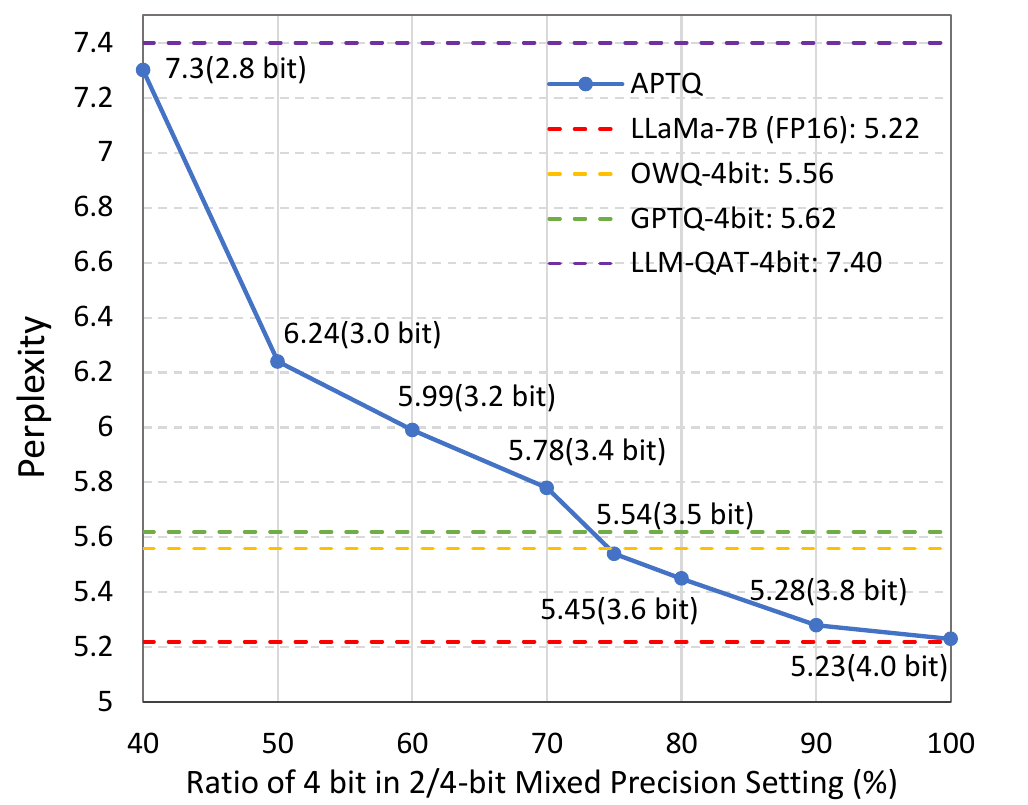}
\caption{Comparative perplexity results of LLaMa-7B using APTQ at various 4-bit ratio against others on C4 dataset}
\label{fig: compare_ratio}
\end{figure}

As depicted in Table~\ref{tab:zero-shot-accuracy}, we benchmark the APTQ framework against current SOTA PTQ methodologies applied to the LLaMa-7B model. Our findings illustrate that APTQ, when configured to 3.8 bits, sustains a remarkably minimal deviation in accuracy, with a diminutive average accuracy drop of only 0.32 points from the full-precision model. Even when the APTQ is optimized down to an average of 3.6 or 3.5 bits, it still consistently outperforms the majority of 4-bit PTQ models. These findings demonstrate that APTQ excels in zero-shot tasks with minimal bit usage, highlighting its effectiveness in deploying large-scale language models in environments with limited computational resources. This underscores APTQ's advantage in resource-efficient performance.

\vspace{-0.2cm}
\subsection{Ablation Study}
Furthermore, we present an ablation study to validate the superiority of APTQ over manual block-wise quantization schemes. Given that quantization is performed on a layer-wise basis, the most intuitive mixed-precision quantization strategy is to uniformly quantize all layers within each block. Here, we compare this conventional approach with APTQ on the LLaMa-7B model tested on the C4 dataset, with perplexity as the evaluation metric. The results in Table~\ref{tab:quantization_comparison} reveal APTQ's efficacy over manual block-wise quantization for LLaMa-7B on C4, reflected in its consistently lower PPL across various quantization ratios.

\begin{table}[t]
\small
\setlength\tabcolsep{2pt} 
\setlength{\belowcaptionskip}{-0.3cm}
\centering
\label{tab:ablation-comparison}
\caption{Ablation Study: Comparison of APTQ and Manual Block-wise Quantization on LLaMa-7B's C4 Perplexity}
\begin{tabular}{lccc}
\toprule
\textbf{Method} & \textbf{Ratio of 4-bit} & \textbf{Avg bit} & \textbf{Perplexity $\downarrow$} \\
\hline
Manual Block-wise & 75\% & 3.5 & 5.84 \\
\textbf{APTQ-75}\% & 75\% & 3.5 & 5.54 \\
Manual Block-wise & 50\% & 3.0 & 7.04 \\
\textbf{APTQ-50}\% & 50\% & 3.0 & 6.24 \\
\bottomrule
\end{tabular}
\vspace{-0.5cm}
\label{tab:quantization_comparison}
\end{table}

% \vspace{-0.2cm}
\section{Conclusion}
\label{sec:conclusion}
This paper presented an Attention-aware Post-Training Mixed-Precision Quantization (APTQ) algorithm for quantizing large language models to mixed precisions. APTQ is a promising post-training quantization strategy by utilizing the second-order information of each layer's weights with consideration of the nonlinear effect of attention outputs. Furthermore, the Hessian trace is developed as a sensitivity measurement to further achieve mixed 2/4-bit precision. 
For LLM LLaMa-7B,  APTQ surpasses previous quantization methods, achieving an average of 4 bits with a 5.22 perplexity, nearly equivalent to full
precision in the C4 dataset. 
Furthermore, under the zero-shot LLM setting, APTQ achieves the state-of-the-art results 68.24\% and 70.48\% accuracy at an average bitwidth of 3.8 for LLaMA-7B and LLaMa-13B, respectively, indicating that APTQ can achieve a deeply quantized solution for large language models without sacrificing accuracy.  

\section{Acknowledgement}
This work was supported by Shenzhen Science and Technology Program (Grant No. KQTD20200820113051096), Science and Technology Innovation Committee Foundation of Shenzhen (Grant No. JCYJ20220818100217038), and by the Theme-based Research Scheme (TRS) project T45-701/22-R, Hong Kong SAR.

\vspace{-0.2cm}
\bibliographystyle{ACM-Reference-Format}
\bibliography{main}

\end{document}